# Evaluation of Morphological Embeddings for the Russian Language

Vitaly Romanov
Innopolis University
Innopolis, Russia
v.romanov@innopolis.ru

Albina Khusainova
Innopolis University
Innopolis, Russia
a.khusainova@innopolis.ru

## ABSTRACT

A number of morphology-based word embedding models were introduced in recent years. However, their evaluation was mostly limited to English, which is known to be a morphologically simple language. In this paper, we explore whether and to what extent incorporating morphology into word embeddings improves performance on downstream NLP tasks, in the case of morphologically rich Russian language. NLP tasks of our choice are POS tagging, Chunking, and NER – for Russian language, all can be mostly solved using only morphology without understanding the semantics of words. Our experiments show that morphology-based embeddings trained with Skipgram objective do not outperform existing embedding model – FastText. Moreover, a more complex, but morphology unaware model, BERT, allows to achieve significantly greater performance on the tasks that presumably require understanding of a word's morphology.

## Keywords
Embeddings evaluation; morphological embeddings; Russian language

## 1. INTRODUCTION

One of the most significant shifts in the area of natural language processing is to the practical use of distributed word representations. Collobert et al. [1] showed that a neural model could achieve state-of-the-art results by relying almost only on the learned word embeddings. In modern language processing architectures, high quality pre-trained representations of words are one of the major factors of the resulting model performance.

Although word embeddings became ubiquitous, there is no single benchmark on evaluating their quality [2], and popular intrinsic evaluation techniques are subject to criticism [3]. Researchers very often rely on intrinsic evaluation, such as semantic similarity or analogy tasks. While intrinsic evaluations are simple to understand and conduct, they do not necessarily imply the quality of embeddings for all possible tasks [4].

In this paper, we concentrate on the evaluation of morphological embeddings for the Russian language. Given its highly fusional morphology and large number of wordforms, we were interested whether training word embeddings based on morphological information has a positive effect on performance in downstream NLP tasks. Over the last decade, many approaches tried to include sub-word information into word representations. Such approaches involve additional techniques that perform segmentation of a word on morphemes [5][6]. The presumption is that we can potentially increase the quality of distributional representations if we incorporate these segmentations into the language model (LM).

Several approaches for including morphology into word embeddings were proposed, but the evaluation often does not compare proposed embedding methodologies with the most popular embedding vectors: Word2Vec [7], FastText [8], GloVe [9]. In this paper, we aim at answering the question of whether morphology-based embeddings can be useful specifically for languages with rich morphology (such as Russian). Our contribution is the following:

1. We evaluate two morphology-based embedding models, comparing them with other popular models as simple as Skip-gram (SG) and as complex as BERT [10] for the Russian language.

2. We test morphological embeddings on several downstream tasks other than language modeling: POS-tagging, Named Entity Recognition, and Chunking.

3. We show that morphology-based embeddings do not outperform simpler FastText [8] model on the chosen downstream tasks.

The rest of the paper is organized as follows. Section 2 contains an overview of existing approaches for morphological embeddings and methods of their evaluation. Section 3 explains embedding models that we have tested. Section 4 explains our evaluation approaches. Section 5 describes the results.

## 2. RELATED WORK

The idea to include sub-word information into word representation is not new, but one needs morphological segmentation of words, which is not easy to obtain. Most of the time, researchers rely on unsupervised morphology mining tools like Morfessor [6].

Morphological language models (LM) use different machine learning approaches starting from Skipgram-like techniques, and ending with multilayer recurrent models. Luong et al. [11] created a hierarchical RNN LM, which performed well on word similarity task. Cao and Rei [12] created Char2Vec BiLSTM LM that generates word embeddings. Their model excels at the syntactic similarity.

Many approaches use simple composition, e.g., sum, of morpheme vectors, to generate word embeddings. Botha and Blunsom [13] were ones of the first to apply morpheme-level composition for log-bilinear LM. They showed a considerable drop in perplexity and also tested their model on word similarity (tested against n-gram language model) and translation task. Qiu et al. [14] incorporated morphemes into CBOW model. Their approach tries to predict morphemes in the current word. They were able to show that including morphemes in word vectors for English language leads to improved metric of semantic similarity. El-kishky et al. [15] develop a LM that jointly learns morpheme segmentations. Their method achieved lower perplexity than FastText and SG.

A slightly different approach was taken by Cotterell [16] who optimized a log-bilinear LM with multitask objective. Unlike in other techniques, they tried to predict a morphological tag, and not morphemes in the word. They tested resulting vector similarity against string distance (morphologically close words have similar substrings) and found that their vectors surpass Word2Vec by a large margin.

Bhatia et al. [17] constructed a hierarchical graphical model that incorporates word morphology that predicts the next word. They compare their model with Word2Vec and the one described by Botha and Blunsom [13]. They found that their method improves results on word similarity but is inferior to approach by Botha and Blunsom [13] in POS-tagging.

Another group of methods tries to incorporate arbitrary morphological information into embedding model. Avraham and Goldberg [18] observe that it is impossible to achieve both high semantic and syntactic similarity on the Hebrew language. Instead of morphemes, they use other linguistic tags for the word, i.e., lemma, the word itself, and morphological tag. Chaudhary et al. [19] took this approach to the next level. Besides including morphological tags, they also include morphemes and character n-grams. Their LM was tested on transfer learning task, where embeddings learned on Turkish language were used in Uighur, and Hindi – in Bengali. They also tested the result on NER and monolingual machine translation.

Another approach that deserves being mentioned here is FastText by Bojanowski et al. [8]. They do not use morphemes explicitly, but instead rely on sub-word character n-grams that store morphological information implicitly. This method achieves high scores on both semantic and syntactic similarities, and by far is the most popular word embedding model that captures word morphology.

In this work, we are going to investigate the performance of word vectors for downstream tasks such as POS-tagging and Chunking. We will use SGNS as the baseline, and compare it with FastText that captures morphology implicitly, and simple LM based on composition of morpheme representations. Finally, we will compare these embeddings with contextual word representations (such as BERT embeddings).

## 3. EMBEDDING TECHNIQUES
### 3.1 Sub-word Units
Morphology-based embeddings proved to be useful for at least one task, i.e., language modeling. Indeed, many prior papers observed the decrease in perplexity after morphology information was introduced into the embeddings. In this paper, we are trying to understand how useful this modeling power can be for more grounded downstream tasks, namely POS-tagging, Chunking, and NER. We are comparing:

1. SGNS
2. FastText
3. Morphemes-based embeddings
4. Morphemes+n-grams-based embeddings which incorporate both character n-grams (FastText) and morphological units

All of these models are trained with the Skip-gram negative sampling objective:

$$\frac{1}{T}\sum_{t=1}^{T}\sum_{-m\leq j\leq m, j\neq 0} \log \sigma\left(s(w_j, w_t)\right) + \sum_{i=1}^{k} \mathbb{E}_{w\sim P_n(w_t)} \log \sigma\left(s(w, w_t)\right)$$

FastText and morpheme-based embeddings make use of sub-word units. The fourth method combines both FastText's character n-grams and morpheme segmentation. We use an identical approach to incorporate sub-word information in all three cases.

$$s(w_j, w_t) = \sum_{s \in S_{w_t}} v_s^T v_{w_j}$$

where $S_{w_t}$ is the set of n-gram segments or morphemes for the current word.

### 3.2 Contextual Embeddings
The general direction of modern NLP is towards contextual representations like ElMO [20] and ULMFiT [21]. While it is very interesting to compare the influence of different types of embeddings on downstream tasks, pre-trained models are hard to obtain for languages other than English. Thankfully, BERT is a multilingual model, and we make use of it in our experiments. We observe in practice that BERT significantly outperforms traditional embedding approaches even on morphologically rich languages.

There is also another family of contextual embeddings based on Deep Semantic Similarity Models (DSSM), but we leave them out of consideration for now.

## 4. EXPERIMENTS AND EVALUATION
To understand the effect of using morphemes for training word embeddings, we performed extrinsic evaluations of SG, FastText, BERT, and introduced earlier Morphemes-based and Morphemes+n-grams-based models for the Russian language. We evaluated these models on POS-tagging, Named Entity Recognition, and Chunking tasks.

### 4.1 Data and Training Details
We used the first 1GB of unpacked Russian Wikipedia dump[1] for training all our models except BERT.

We used our own code to train all models except BERT. The goal at this stage is to capture tendencies in model behavior. We assume that even though the training data is relatively small, we will be able to see the general differences between embedding approaches.

Training parameters for all regular embeddings are picked the same and similar to the original values used by Mikolov et al. [22] including learning rate, subsampling parameters, and negative sampling distribution. Vocabulary size was constrained to 100 000 words. Morphemes for Russian languages were obtained using an open source segmentation tool[2].

We will refer to FastText as FT, Morphemes-based embeddings as Morph, and Morphemes+n-grams-based as MorphNG in the following discussion.

---

[1] https://dumps.wikimedia.org

[2] https://github.com/kpopov94/morpheme_seq2seq

## 4.2 Neural Architecture Overview

### 4.2.1 Architecture for embeddings

Embeddings are the driver of neural NLP. Collobert et al. [1] showed that state-of-the-art results are possible to achieve with unsupervised feature learning for text processing applications. They introduced a unified architecture for POS-tagging, Chunking, NER and other. We employ a similar approach and use convolutional NN for token level classification. The architecture consists of four components:

1. Embedding matrix. We do not use any additional discrete features in our implementation other than words themselves because the primary goal is to see how word embeddings encode relevant features.
2. Convolutional layer with non-linearity. Unlike contextual embeddings, traditional word embeddings exist out of context, and convolutional layer allows to capture the interaction between neighboring words.
3. Dense layer with non-linearity. The goal of this layer is to extract additional features.
4. CRF layer. The main purpose of this layer is to model the conditional distribution of the tags.

We use this model for POS, Chunking and NER for Skip-gram and sub-word unit-based embeddings.

### 4.2.2 Architecture for Contextual Embeddings

Unlike regular embeddings, BERT is based on a bidirectional model and inherently captures information about neighboring words. For this reason, there is no need for a convolutional layer. Also, BERT provides a multilayer view of the data, where each layer can potentially model different aspects of the input. We use the simple approach to utilize the multilayer view of the data—we sum embeddings from all the layers of BERT encoder. The rest of the architecture consists of two consecutive layers interacting with a token embedding produced by BERT. Devlin et al. [10] provide a hint on how to implement a tagging algorithm using BERT. They did not use any convolutional or CRF layer, and the classification decision was made solely based on the single token embedding. The trick is that BERT does sub-word segmentation. Devlin et al. [10] suggest to do the classification of only the first segment of the word, and exclude the additional segments like these:

| Andy  | Hu    | ##eman | was | a | de | ##cent | person |
|-------|-------|--------|-----|---|----|--------|--------|
| I-PER | I-PER | X      | O   | O | O  | X      | O      |

Here the segments that begin with ## do not participate in the optimization of the objective.

In all experiments, we used "frozen" embeddings, meaning we did not train the embedding layer.

It is worth noting here that BERT's sub-word segmentation is not equivalent to morphological segmentation, which is demonstrated by following example. The compound Russian word авиаракетостроение should be segmented into morphemes as авиа|ракет|о|строени|е, but the BERT's tokenizer will produce а|ви|ара|кет|ост|рое|ние. BERT's segmentation is based on char-n-gram frequencies, which does not always correspond to proper morphemes.

Further, we are going to describe the details of our downstream tasks.

## 4.3 POS-Tagging

For the task of Part-of-Speech tagging we use a Russian language dataset with detailed morphological tags and chunking markers[3]. This dataset is a subset of SynTagRus [23]. It contains 49 136 sentences and 458 unique POS tags. Due to time constraint, we train models only for 20 epochs and report the best result. The quality of POS-tagging is measured with accuracy. The results are given in Table 1.

We see that in this task SG performs worse than other models, and FT shares the first place with MorphNG.

## 4.4 Chunking

The same dataset that we used for POS-tagging contains chunking information. We use identical neural architecture to label each token in the sentence. There is no significant class imbalance, therefore, we use per word accuracy to measure the quality on the task.

The same pattern as with POS-tagging task is observed here, see Table 2. FT, Morph and MorphNG perform on the same level, leaving SG behind.

**Table 1. Accuracy on POS-tagging and Chunking tasks, F1-score for Named Entity Recognition task for models trained on 1GB Russian text**

|          | SG   | FT   | Morph | MorphNG |
|----------|------|------|-------|---------|
| POS      | 0.84 | **0.88** | 0.87 | **0.88** |
| Chunking | 0.82 | **0.84** | **0.84** | 0.83 |
| NER      | **0.94** | **0.94** | **0.94** | **0.94** |

## 4.5 Named Entity Recognition

We used Named_Entities_3 collection introduced in [24]. Unlike others, this dataset uses BILUO tagging scheme. Due to significant disbalance between classes, we use F1 score as the quality metric.

As we see (Table 1), there is no clear difference between models on this task. All the models show decent performance on NER task which probably suggests that the dataset is easy to learn.

Overall, FT, Morph, and MorphNG show similar performance on all tasks, and we don't observe clear benefits of morphology-based models over FastText, which is also cheaper to train and needs less supervision.

We also experimented with continuing training embeddings for the tasks instead of providing fixed embeddings. In this scenario, all models were able to learn and reach the same level of accuracy/F1-score for all tasks.

## 4.6 Comparison Fairness

It is unfair to directly compare BERT with other methods because it was trained on a significantly larger dataset[4]. We observe that on most of the tasks for Russian language FastText outperformed sibling approaches. For this reason, we used an official pretrained version of FastText vectors[5], that were trained on a significantly larger dataset than ours, see Table 2. From the comparison, we can see that BERT significantly outperforms FastText. This can simply be explained by a much higher modeling power of bidirectional transformer architecture.

---

[3] http://web-corpora.net/wsgi/chunker.wsgi/npchunker/npchunker/

[4] https://github.com/google-research/bert

[5] https://fasttext.cc

**Table 2. Accuracy on POS-tagging and Chunking tasks, F1-score for Named Entity Recognition task for models trained on large datasets of Russian texts**

|  | **BERT** | **FT big** |
|---|---|---|
| POS | **0.96** | 0.91 |
| Chunking | **0.95** | 0.90 |
| NER | **0.96** | 0.95 |

## 5. RESULTS

In this paper, we compared five word embedding approaches using a set of downstream tasks for the Russian language. One of the methods is the classical Skip-gram with negative sampling, three of them are the extensions of Skip-gram objective that allow incorporating sub-word information into word vectors, such as character n-grams or a set of morphemes. The main inquiry was about the benefits of providing morphological information to word embeddings. We wanted to see how this additional information helps us in downstream tasks. Experiments showed that morphology-based embeddings exhibit feature encoding qualities akin to FastText, which suggests that for the chosen tasks the choice of sub-word units—morphemes or n-grams doesn't make much difference. At the same time, it is harder to do a good segmentation of words into morphemes than into character n-grams and additional models are required. The primary motivation for using morphemes for word embeddings in morphologically rich languages is that we can encode more semantic information. Since there is no strong support in favor of using morphemes, we suggest that FastText is the best choice among Skip-gram-based embeddings for the NLP tasks discussed above for the Russian language. However, BERT showed superior performance on all tasks, and it suggests that contextuality brings more benefit than modeling of sub-word units. The interesting detail is that models fed with BERT embeddings were able to learn only when the learning rate was lowered to 1e-4. Despite this model being more superior in terms of the quality of the results, BERT is very resource hungry, and so, not suitable for every setting.